\documentclass[10pt]{article}
\usepackage[a4paper, total={6.2in, 10in}]{geometry}
\usepackage{multicol} 
\usepackage{graphicx} 
\usepackage{times}
\graphicspath{ {./images/} }
\usepackage{natbib}
\setcitestyle{authoryear,open={(},close={)}} 

\title{\textbf{The Role of Text-to-Image Models in Advanced Style Transfer Applications:} A Case Study with DALL·E 3}
\author{
    {Ike Ebubechukwu$^1$} \\
    \texttt{ebubechukwu.ike@studenti.unitn.it}
}

\date{
    \textbf{{A University of Trento Project}} \\
    {Domain adaptation}
}
\begin{document}
\maketitle
\begin{center}
    \includegraphics[width=15cm]{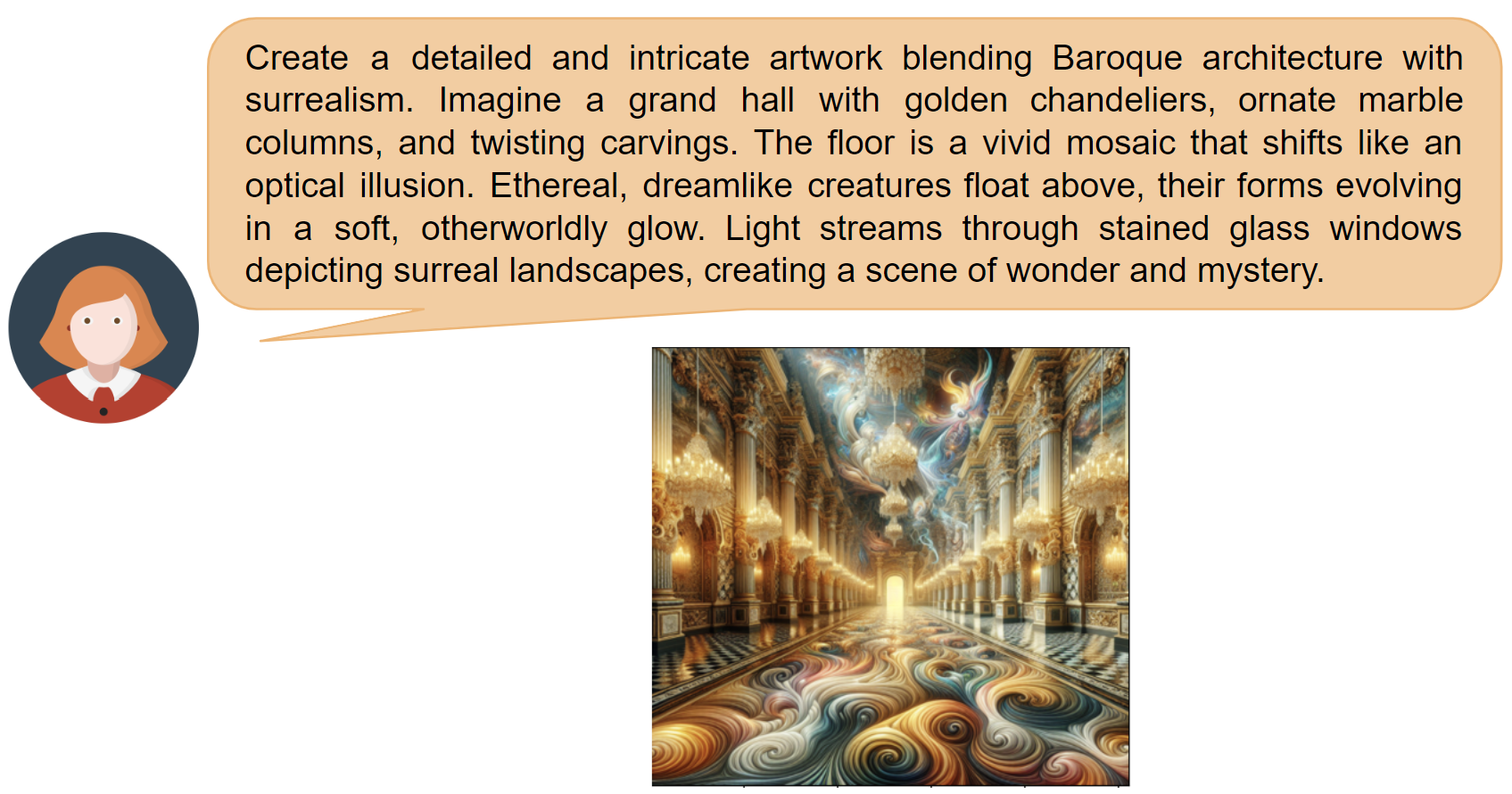}
\end{center}
\noindent \small \textbf{Figure 1:} Shows the prompt given to DALL·E 3 to create an intricate artwork blending Baroque architecture with surrealism. The resulting image has a resolution of 1024x1024 pixels, showcasing a grand hall filled with golden chandeliers, ornate marble columns, and vivid mosaics, embodying the detailed and imaginative style that DALL·E 3 can achieve. \newline

\begin{abstract}
\noindent While DALL·E 3 has gained popularity for its ability to generate creative and complex images from textual descriptions, its application in the domain of style transfer remains slightly underexplored. This project investigates the integration of DALL·E 3 with traditional neural style transfer techniques to assess the impact of generated style images on the quality of the final output. DALL·E 3 was employed to generate style images based on the descriptions provided and combine these with the Magenta Arbitrary Image Stylization model. This integration is evaluated through metrics such as the Structural Similarity Index Measure (SSIM) and Peak Signal-to-Noise Ratio (PSNR), as well as processing time assessments. The findings reveal that DALL·E 3 significantly enhances the diversity and artistic quality of stylized images. Although this improvement comes with a slight increase in style transfer time, the data shows that this trade-off is worthwhile because the overall processing time with DALL·E 3 is about \textbf{2.5 seconds} faster than traditional methods, making it both an efficient and visually superior option. \newline
\end{abstract}

\begin{multicols}{2}
\section{Introduction}
The emergence of advanced AI models like DALL·E 3 has revolutionized the landscape of image generation and style transfer. DALL·E 3, developed by OpenAI \citep{betker2023improving}, is a powerful model capable of generating highly detailed and diverse images based on textual descriptions \citep{reddy2021dall}. This ability to seamlessly translate language into visual representations opens up new possibilities for creative applications, particularly in the field of artistic style transfer \citep{psychogyios2023samstyler}.\newline
\noindent Artistic style transfer, a technique that applies the visual style of one image to the content of another, has been an area of active research and development since the pioneering work of Gatys et al., 2016. Traditional methods primarily focused on transferring basic visual elements like color and texture, often overlooking the more intricate aspects of an artist’s style, such as brushstrokes, mood, and composition. These earlier approaches, while effective to a degree, were limited in their ability to capture the full essence of an artist’s unique expression \citep{xu2023stylerdalle}. \newline
\noindent DALL·E 3 addresses these limitations by leveraging its vast training on large-scale datasets that include both images and textual descriptions. This model is not only capable of understanding and generating visual content but also infuses that content with the specific stylistic nuances described in text \citep{betker2023improving}. When combined with neural style transfer techniques, DALL·E 3 enhances the stylistic diversity and fidelity of the output images, making it possible to apply complex, abstract styles that were previously difficult to achieve. \newline

\includegraphics[width=7cm]{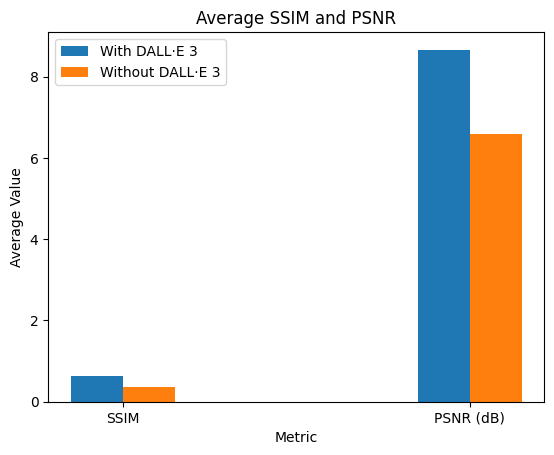}
\small \textbf{Figure 2:} Comparison of Average SSIM and PSNR between Style Transfer Models with and without DALL·E 3 generated image. The results show that images generated with DALL.E 3 outperform using already existing images, in maintaining structural similarity (SSIM) with the original content image, while still maintaining a way better quality (PSNR) after the style transfer is performed. \newline
    
\noindent In this project, DALL·E 3 was integrated with the Magenta Arbitrary Image Stylization model—a neural network designed for real-time artistic style transfer (Ghiasi et al., 2017). This combination allows users to input a textual description of a desired style, which DALL·E 3 then transforms into a corresponding style image. The neural style transfer model applies this style to a content image, resulting in a final image that maintains the content of the original while adopting the intricate stylistic features generated by DALL·E 3.\newline
\noindent The Magenta model \citep{ghiasi2017exploring}, operates by separating the content and style elements of an image and then recombining them in a way that preserves the structural integrity of the content while overlaying the stylistic features of another image (see \textbf{Figure 5} for model architecture). This model is highly versatile and has been optimized for arbitrary style transfer, meaning it can adapt to a wide range of styles without requiring retraining. Its integration with DALL·E 3 introduces a new dimension of flexibility and creativity, as users can now define styles in natural language, unlocking a broader array of artistic possibilities.\newline
\noindent Through this integration, the aim is to explore the potential of using advanced text-to-image models like DALL·E 3 in the domain of artistic style transfer, pushing the boundaries of what can be achieved in terms of creativity, user engagement, and the quality of the stylized outputs. This approach should not only enhance the visual quality of the images but also democratize the creative process, allowing even those without technical expertise to produce highly stylized artwork with ease. To conclude, the main contribution:
\begin{itemize}
    \item  Integrating DALL·E 3, a powerful text-to-image generation model, with traditional neural style transfer techniques, which offers more personalized and diverse artistic results.
    
    \item Analyzing the trade-offs between using DALL·E 3 for style image generation and conventional style transfer methods. By quantitatively assessing the impact on processing time, SSIM, and PSNR.
\end{itemize}

\section{Background}
As discussed earlier, Neural style transfer has been extensively explored since \citep{gatys2016image} introduced a method leveraging convolutional neural networks (CNNs) to extract and transfer artistic styles between images. Subsequent works have refined this approach by incorporating diversified styles (\citealt{ulyanov2017improved}; \citealt{wang2020diversified}) and attention mechanisms (\citealt{park2019arbitrary}; \citealt{liu2021adaattn}; \citealt{yao2019attention}). \cite{chen2021artistic} and \cite{wang2022aesust}, further advanced the field by applying generative adversarial networks (GANs) to bridge the gap between AI-generated and human-created artworks. This study builds upon these by integrating DALL·E 3 for text-driven style generation, eliminating a constant need for specific datasets. Additionally, large-scale text-to-image models, such as those by \citep{ramesh2021zero} as discussed in the third week of our class on language and vision, have shown remarkable capabilities in generating diverse images from textual inputs. 

\noindent This aligns with the approach, which leverages DALL·E 3's generative power and the magenta model to enhance the diversity and quality of style transfer, grounding it in visual-language representations. Building on these developments, this research aims to address the following questions:
\begin{enumerate}
    \item Does integrating DALL·E 3 with traditional neural style transfer models improve the quality and diversity of stylized images?
    
    \item What are the qualitative differences in user experience and satisfaction when interacting with a style transfer system that incorporates DALL·E 3?
\end{enumerate}
\end{multicols}

\begin{center}
    \includegraphics[width=15cm]{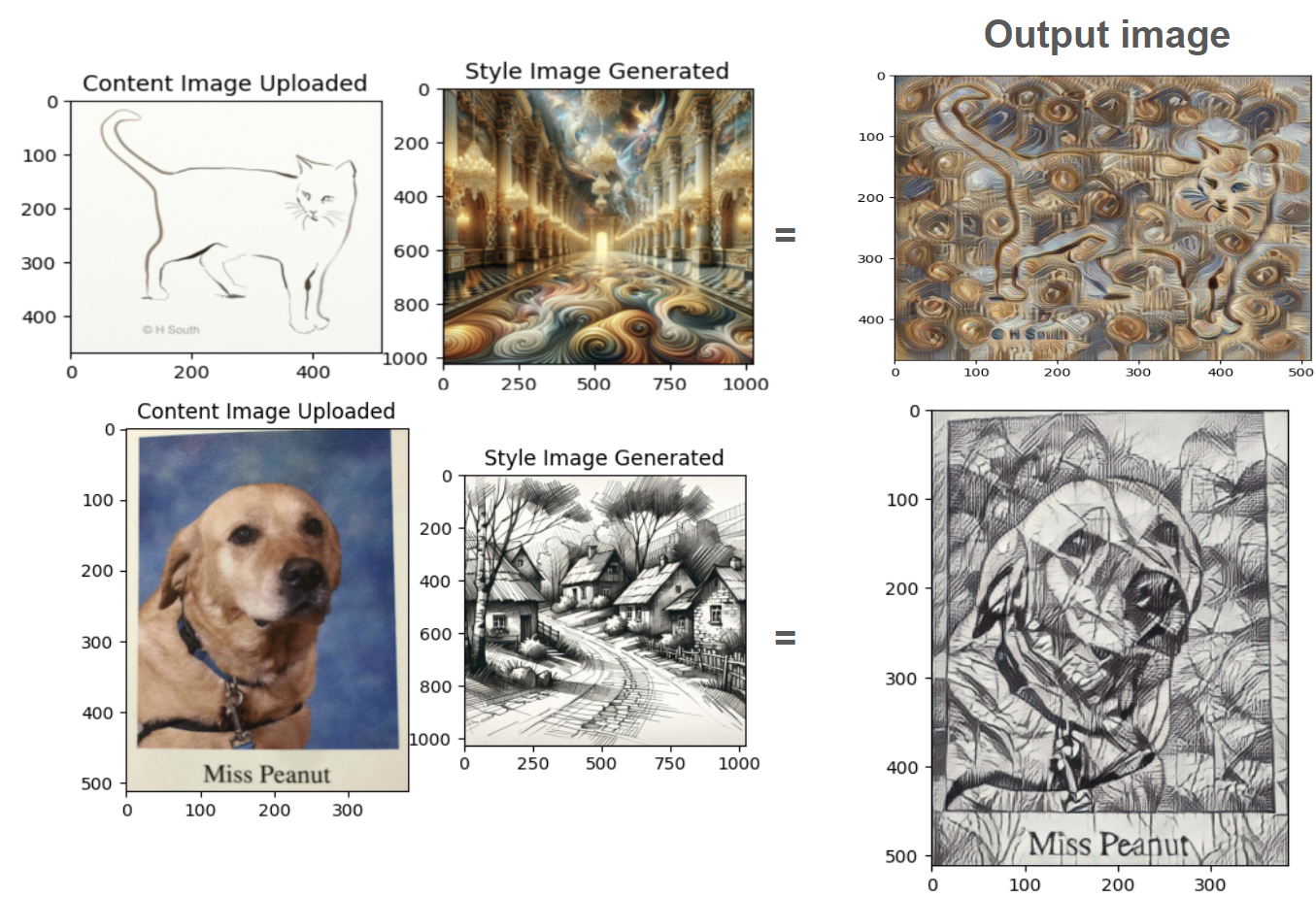}
\end{center}
\noindent \small \textbf{Figure 3:} Style transfer using DALL·E 3 generated images across different domains. The top row combines a simple sketch of a cat with a Baroque-inspired style, while the bottom row blends a dog photo with a pencil sketch style. Both content images are from the Sketch and Real datasets of DomainNet \citep{peng2019moment}.

\begin{multicols}{2}
\section{Experimental Setup}
\subsection{Datasets:}For this project, a minimal dataset was utilized to test the integration of DALL·E 3 with a neural style transfer model. Each content image was selected from three distinct domains: Real, Sketch, and Painting, within the DomainNet dataset put together by \cite{peng2019moment}. These domains provide a diverse set of visual characteristics, allowing for a thorough evaluation of the style transfer process. Additionally, a modern art image was included from the "Abstract Portrait, Wall Art | Canvas Prints by F. Abderrahim" collection. No preprocessing was performed on all four images, they were used as-is to maintain the integrity of their original domains.
\subsection{DALL·E 3 Architecture:} DALL·E 3, developed by OpenAI, is a generative model that converts textual descriptions into high-quality images. The model architecture is based on transformers, specifically using a decoder-only architecture. Mathematically, DALL·E 3 models the probability distribution of images given textual descriptions using an autoregressive approach: \[ P(I | T) = \prod_{i=1}^{n} P(I_i | I_{1:i-1}, T) \]
\textbf{Where:} \(I\) represents the image, \(T\) the textual description, and \(I_i\) the image token at position \(i\). \newline
\noindent The model is trained on a vast dataset of text-image pairs, allowing it to generate highly detailed and contextually relevant images. DALL·E 3 was selected because it is currently the best model in generating highly detailed and diverse images from textual descriptions\newline
\end{multicols}
\begin{center}
    \includegraphics[width=15cm]{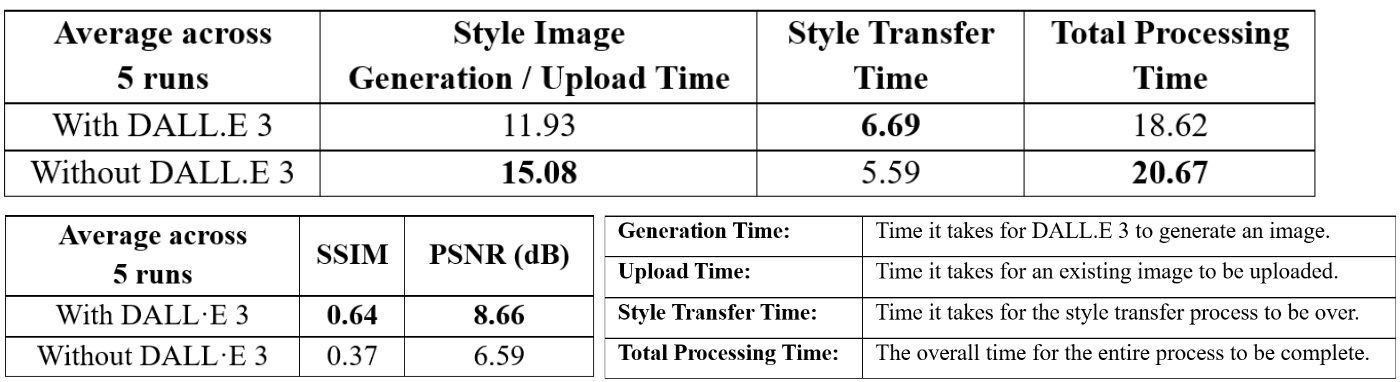}
\end{center}
\noindent \small \textbf{Figure 4:} Comparison of average processing time, SSIM, and PSNR across five runs with and without DALL·E 3. The numbers in bold signify the highest value in each respective column, indicating the superior performance in those metrics. On the right, the table explains what each type of time represents in the process.

\begin{center}
    \includegraphics[width=15cm]{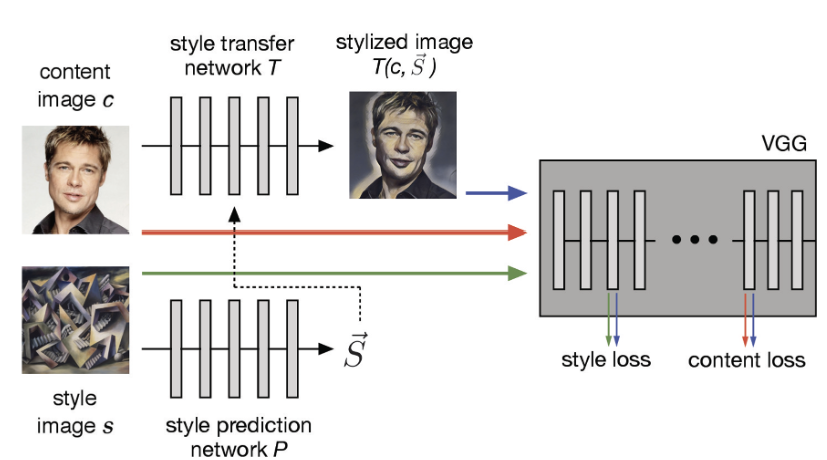}
\end{center}
\noindent \small \textbf{Figure 5:} Diagram of the style transfer model architecture from \citet{ghiasi2017exploring}. The style prediction network (P) generates an embedding vector from the input style image, which then guides the style transfer network to transform the photograph into a stylized image. Content and style losses are calculated using the VGG network \citep{gatys2015neural}, and the architecture is based on \citet{dumoulin2016learned} with the style prediction network following Inception-v3 \citep{szegedy2016rethinking}.
\begin{multicols}{2}
\subsection{Style Transfer Model Architecture:} For the style transfer, the "Magenta Arbitrary Image Stylization" model available on TensorFlow Hub was utilized. This model is built upon the work of Ghiasi et al. (2017) and is designed to apply arbitrary styles to content images in real-time. The model consists of two main components: the style prediction network and the transformation network. \newline
\noindent Mathematically, the style transfer process can be described as minimizing the following loss function:
\[ \mathcal{L}_{total} = \alpha \mathcal{L}_{content}(C, S) + \beta \mathcal{L}_{style}(C, S) \] \newline
\noindent \textbf{Where:} \( \mathcal{L}_{content}(C, S) \) measures the difference between the content representation of the content image \(C\) and the stylized image \(S\). \( \mathcal{L}_{style}(C, S) \) measures the difference in style between the style image and the stylized image. \(\alpha\) and \(\beta\) are weights that balance content and style preservation. \newline
The Magenta model was chosen due to its capability to apply arbitrary styles to content images in real-time \citep{ghiasi2017exploring}. The model doesn't require both images to be the same size, which was important in preserving the quality of the generated images. Also, its proven efficiency in handling various artistic styles made it an ideal choice for experimenting with the styles generated by DALL·E 3.
\subsection{Combining DALL·E 3 and Style Transfer:} To achieve the final stylized images, DALL·E 3 was first used to generate style images based on specific textual prompt (e.g \textit{Make the image look like a simple modern art woman's face}.) These generated style images were then fed into the Magenta model as style references. The Magenta model then applied these styles to the content images, as seen in Figures 3 and 6.
\end{multicols}

\begin{center}
    \includegraphics[width=13cm]{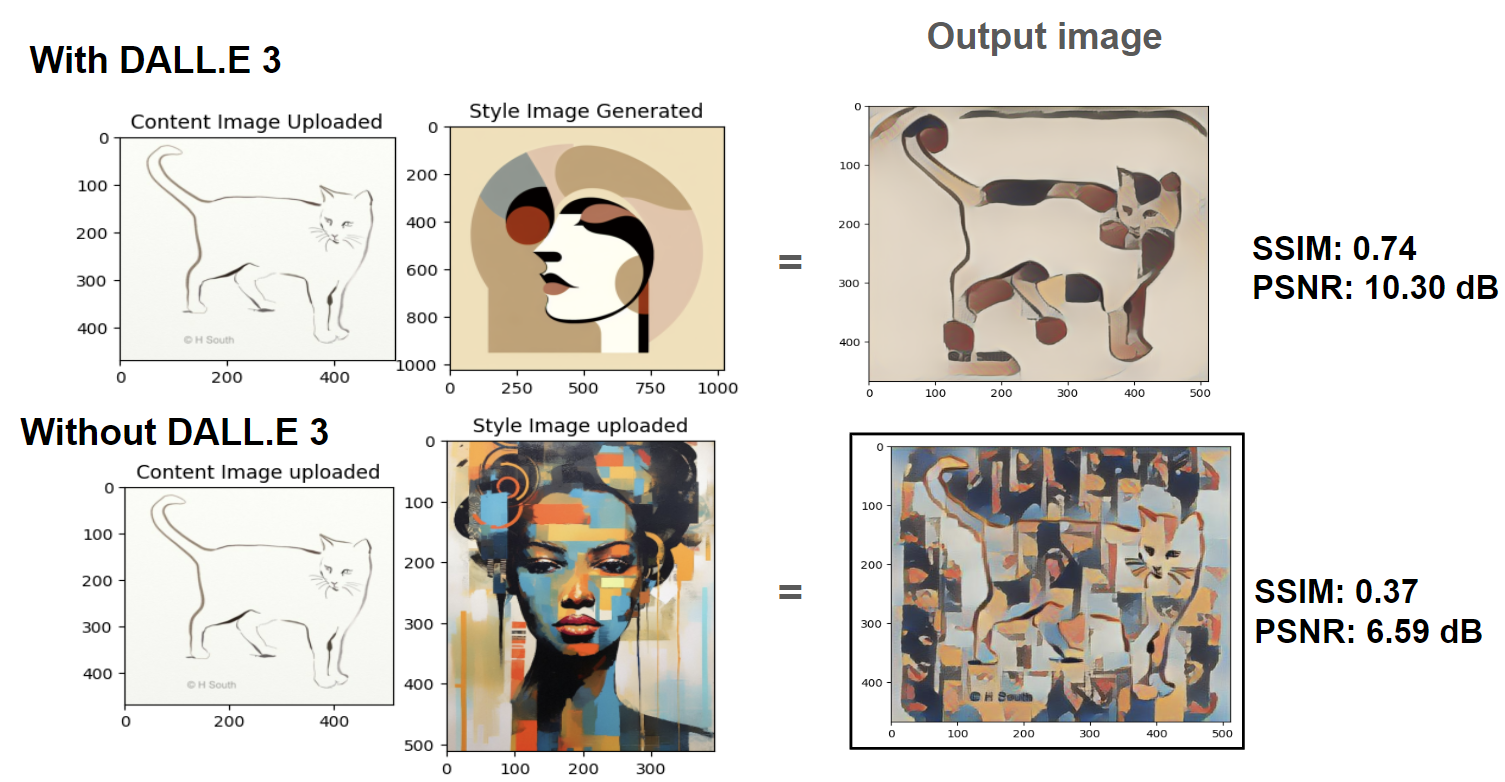}
\end{center}

\noindent \small \textbf{Figure 6:} Comparison of style transfer results using the same modern art image as the content image, with and without DALL·E 3. The top row, using DALL·E 3, shows a 50\% higher SSIM and 30\% higher PSNR, indicating better quality and structural similarity. The bottom row, without DALL·E 3, displays noticeably lower SSIM and PSNR values.

\begin{multicols}{2}
\subsection{\textbf{Metrics:}}
\subsubsection{Structural Similarity Index Measure (SSIM)} SSIM is a widely used metric that measures the similarity between two images by comparing their luminance, contrast, and structure \citep{nilsson2020understandingssim}. The SSIM between two images \( x \) and \( y \) is calculated as follows:
\[ SSIM(x, y) = \frac{(2\mu_x \mu_y + C_1)(2\sigma_{xy} + C_2)}{(\mu_x^2 + \mu_y^2 + C_1)(\sigma_x^2 + \sigma_y^2 + C_2)} \]
Where: \( \mu_x \) and \( \mu_y \) are the mean intensities of images \( x \) and \( y \). \( \sigma_x^2 \) and \( \sigma_y^2 \) are the variances of images \( x \) and \( y \). \( \sigma_{xy} \) is the covariance of images \( x \) and \( y \). \( C_1 \) and \( C_2 \) are constants to stabilize the division with weak denominators.
\subsubsection{Peak Signal-to-Noise Ratio (PSNR)}  PSNR measures the ratio between the maximum possible power of a signal (in this case, the image) and the power of noise that affects its fidelity. It is typically used to quantify the quality of a reconstructed image, with higher values indicating better quality \citep{kelecs2021computation}. The PSNR between two images \( x \) and \( y \) is calculated using the Mean Squared Error (MSE) as follows:
\[ PSNR(x, y) = 10 \cdot \log_{10} \left(\frac{MAX_I^2}{MSE}\right) \]
\noindent Where: \( MAX_I \) is the maximum possible pixel value of the image. For an 8-bit image, this value is 255. \( MSE \) is the Mean Squared Error between the two images, defined as: \[ MSE = \frac{1}{mn} \sum_{i=0}^{m-1} \sum_{j=0}^{n-1} \left[ x(i,j) - y(i,j) \right]^2\] Here, \( m \) and \( n \) are the dimensions of the images \( x \) and \( y \). \newline

\noindent Both metrics were selected because they together offer a comprehensive evaluation of both the perceptual quality and fidelity of the stylized images, which are critical aspects in the context of neural style transfer. 
\section{Results}
\subsection{Quality of the style-transferred Images:} The results demonstrate a clear improvement in image quality with the inclusion of DALL·E 3. Specifically, the average \textbf{SSIM} for images generated \textbf{with DALL·E 3 was 0.64}, compared to \textbf{0.37 without DALL.E 3}. Similarly, \textbf{PSNR} values averaged \textbf{8.66 dB with DALL·E 3}, versus \textbf{6.59 dB without}. These metrics suggest that DALL·E 3 significantly enhances both the structural integrity and overall clarity of the stylized images. \newline
\noindent Moreover, because DALL·E 3 generates different style images for each run, it introduces a wide variety of artistic styles, resulting in the diversity of the output images. This diversity will be crucial for creative applications where uniqueness and variability are highly valued (e.g creating a company logo etc.). These results answer the first research question that integrating DALL·E 3 with traditional neural style transfer models improves the quality and diversity of stylized images. 
\subsection{User Experience and Satisfaction:} The second research question examined how user experience and satisfaction change when using a style transfer model with DALL·E 3 (Generation Time, Upload time, Style Transfer time, and Overall processing time), key factors in user experience, were measured over five runs to get average values. The results showed that with DALL·E 3 the overall processing time showed an average of \textbf{18.62 seconds} compared to the overall processing time without DALL.E 3 which was \textbf{20.67 seconds}. While using DALL·E 3 took on average \textbf{1.1 seconds} longer during style transfer, this is outweighed by the creative freedom it offers, allowing users to explore a wider range of styles and achieve more personalized outcomes.  A possible reason for the Longer Style Transfer Time with DALL·E 3 could be that DALL.E 3 generates higher-resolution images, which take more time to process during style transfer.
\section{Conclusion}
The results of this study demonstrate that integrating DALL·E 3 into traditional neural style transfer significantly enhances both the quality of stylized images and the overall user experience. The improved diversity and artistic quality justify the slight increase in style transfer time. However, this approach does have its limitations:
\begin{enumerate}
    \item The time required to generate images with DALL·E 3, as opposed to using pre-existing images, can become an issue when processing large batches of images.
    \item The Style Transfer Model tends to be slower when working with the high-resolution images generated by DALL·E 3.
\end{enumerate}
To address these challenges, future experiments could include:
\begin{enumerate}
    \item Experimenting with the use of pre-generated style images from DALL·E 3 for batch processing to reduce on-demand generation time.
    \item Exploring the impact of reducing image resolution on processing speed and image quality, or improving the model's ability to handle high-resolution images more efficiently.
\end{enumerate}
By addressing these limitations, the integration of DALL·E 3 into style transfer workflows can be further optimized, making it even more effective for a wide range of creative applications.
\end{multicols}

\clearpage
\bibliography{The_Role_of_Text-to-Image_Models_in_Advanced_Style_Transfer_Applications__A_Case_Study_with_DALL__E_3}
\section{Appendix}

\begin{center}
    \includegraphics[width=15cm]{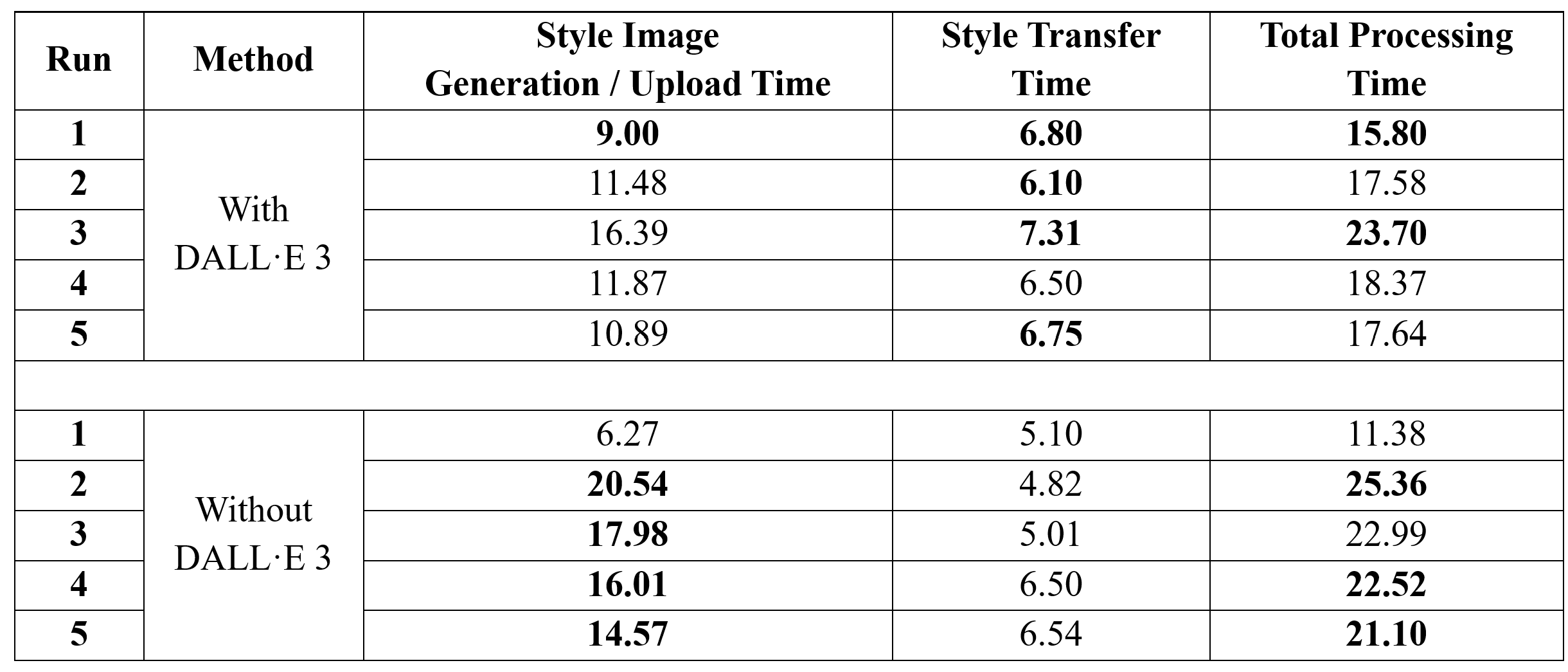}
\end{center}

\noindent \small \textbf{Figure 7:} This table compares the style image generation/upload time, style transfer time, and total processing time for five runs using DALL·E 3 and without DALL·E 3. The bold numbers indicate the highest times recorded in each row.\newline  

\noindent For context, this experiment involved generating a modern art image of a woman's face to match an existing image from the "Abstract Portrait, Wall Art | Canvas Prints by F. Abderrahim" collection. In the "Without DALL·E 3" runs, the same image was used each time, while in the "With DALL·E 3" runs, a different image was generated each time. This table illustrates the time differences between using pre-existing images versus generating new ones with DALL·E 3. Despite the variation in image generation, DALL·E 3 maintained a comparable total processing time, highlighting its efficiency even when generating diverse style images.

\begin{center}
    \includegraphics[width=13cm]{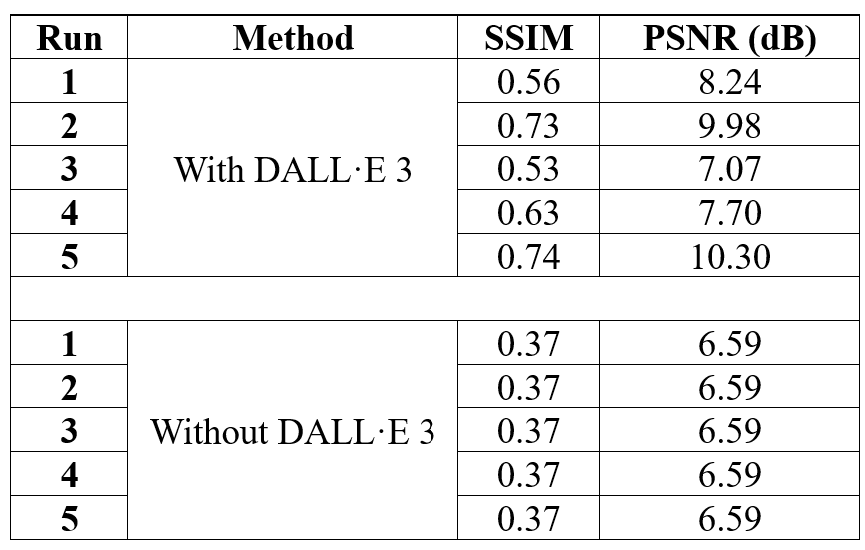}
\end{center}

\noindent \textbf{Figure 8:} This table presents the SSIM and PSNR results from five experimental runs of style transfer with and without DALL·E 3. As shown, the values for the runs without DALL·E 3 remain consistent across all five runs, while the runs with DALL·E 3, which generates a new style image each time, show variability but consistently higher SSIM and PSNR values.
\end{document}